\documentclass{article}

\usepackage{microtype}
\usepackage{graphicx}
\usepackage{subfigure}
\usepackage{booktabs} 
\usepackage{algorithm}
\usepackage{algpseudocode}
\usepackage{xcolor}

\usepackage{authblk}

\usepackage{hyperref}

\usepackage[accepted]{icml2025}

\usepackage{amsmath}
\usepackage{amssymb}
\usepackage{mathtools}
\usepackage{amsthm}
\usepackage{multirow}

\usepackage[capitalize,noabbrev]{cleveref}

\theoremstyle{plain}

\theoremstyle{definition}

\theoremstyle{remark}

\usepackage[textsize=tiny]{todonotes}

\begin{document}

\twocolumn[
\icmltitle{QoS-Efficient Serving of Multiple Mixture-of-Expert LLMs Using Partial Runtime Reconfiguration}

\begin{icmlauthorlist}
\icmlauthor{HamidReza Imani}{yyy}
\icmlauthor{Jiaxin Peng}{yyy}
\icmlauthor{Peiman Mohseni}{yyyy}
\icmlauthor{Abdolah Amirany}{yyy}
\icmlauthor{Tarek El-Ghazawi}{yyy}

\end{icmlauthorlist}

\begin{center}
    \url{https://github.com/hamid-74/Multi-MoE}
\end{center}
\icmlaffiliation{yyy}{Department of Electrical and Computer Engineering, The George Washington University}
\icmlaffiliation{yyyy}{Computer Science and Engineering Department, Texas A\&M University}

\icmlcorrespondingauthor{HamidReza Imani}{hamidreza@gwu.edu}

\icmlkeywords{Machine Learning, ICML}

\vskip 0.3in

]

\printAffiliationsAndNotice{}




\begin{abstract}

The deployment of mixture-of-experts (MoE) large language models (LLMs) presents significant challenges due to their high memory demands. These challenges become even more pronounced in multi-tenant environments, where shared resources must accommodate multiple models, limiting the effectiveness of conventional virtualization techniques. This paper addresses the problem of efficiently serving multiple fine-tuned MoE-LLMs on a single-GPU. We propose a serving system that employs \textit{similarity-based expert consolidation} to reduce the overall memory footprint by sharing similar experts across models. To ensure output quality, we introduce \textit{runtime partial reconfiguration}, dynamically replacing non-expert layers when processing requests from different models. As a result, our approach achieves a competitive output quality while maintaining throughput comparable to serving a single model while incurring a negligible increase in time-to-first-token (TTFT). Experiments on a server with a single NVIDIA A100 GPU (80GB) using Mixtral-8x7B models demonstrate an 85\% average reduction in turnaround time compared to NVIDIA's multi-instance GPU (MIG). Furthermore, experiments on Google's Switch Transformer Base-8 model with up to four variants demonstrate the scalability and resilience of our approach in maintaining output quality compared to other model merging baselines, highlighting its effectiveness.

\end{abstract}


\section{Introduction}

Incorporating the Mixture-of-Experts concept \cite{jacobs1991adaptive, jordan1994hierarchical} into transformer-based language models \cite{he2024instruction} has enabled an expansion in model parameters, improving the quality of generated outputs across various language tasks \cite{shazeer2017outrageously, shazeer2018mesh, fedus2022switch, du2022glam, kim2021scalable, zuo2021taming, lin2021m6, rajbhandari2022deepspeed}. Rather than using a single block (e.g., a feed-forward layer), MoE-based models employ a gating system and multiple parallel instances of the block (known as experts) with identical structures. During the forward pass calculation, based on the current input, the gating system assigns weights to each expert, and one or more top experts will be selected to produce the block's output. This will allow the model to scale in size while imposing negligible computation overhead (in the case of selecting only one expert).

However, such scaling comes with its own drawbacks. MoE LLMs have significantly higher memory footprints and do not fit in most of the recently developed GPU memories. For example, Google's Switch Transformer \cite{fedus2022switch} model has a total of 1.6 trillion parameters which occupies 3.1 TB of memory (16 bits for each parameter). Therefore, for deployment in constrained environments, the model parameters should be partitioned between CPU and GPU memories. Accordingly, in the calculation of forward pass, repetitive host-to-device copy operations will be required which stall the inference pipeline and limit the overall throughput of the system.

At the same time, an increasing number of private entities are adopting customized MoE-based LLMs to meet their specific needs and ensure the required output quality. These LLMs are typically fine-tuned versions of a general-purpose model, trained on specific datasets \cite{jiang2024mixtral} which should be deployed in local settings for efficiency and customization purposes. However, most of these entities lack access to sufficient GPU memory to accommodate large models, as they typically have only a limited number of GPUs. This issue becomes even more challenging when multiple entities attempt to perform inference with their customized models simultaneously in the described constrained environment and must share resources.

Current research efforts to improve MoE inference efficiency and performance focus on reducing the overall memory footprint \cite{lin2023awq, xiao2023smoothquant, huang2024billm, eliseev2023fast, dettmers2023case} and host-to-device communication overhead \cite{kamahori2024fiddler, eliseev2023fast, xue2024moe, soosis} of a single MoE model. Quantization and mixed precision approaches directly reduce the memory footprint which allow the server to fit more parameters on the GPU's memory. In another direction, previously proposed expert caching mechanisms identify the most frequently used experts and prioritize them for loading into GPU memory. As a result, these approaches reduce the communication overhead by increasing the probability of future required experts being available on GPU. However, these approaches face considerable performance degradation in the case of a system that supports the inference of multiple MoEs.

Conventional virtualization and sharing techniques, such as space and time sharing \cite{el2009exploiting, li2011gpu, huang2010reconfiguration}, are also inefficient for this class of problem. Space-sharing approaches, like NVIDIA Multi-Instance GPU (MIG) \cite{choquette2021nvidia}, divide the available GPU resources among the requesting users, which amplifies the communication overhead. In contrast, in time sharing approaches, requests for different models cause the currently loaded model to be offloaded and another model to be uploaded onto the GPU, a process that can take up to two minutes.

In this paper, to optimize the performance and availability of a multi-model inference system, we propose a serving system which is expected to achieve a throughput comparable to serving a single model. This feature will be supported at the cost of a slight increase in time-to-first-token (TTFT) and decrease in generated output quality. Although this paper focuses specifically on fine-tuned LLMs with identical architectures and text generation tasks, the proposed technique is applicable to MoEs with different sizes and number of parameters. In summary, we make the following contributions:

\begin{itemize}
    \item Expert Loading on GPU via Similarity-Based Consolidation: We propose to find a middle point between different fine-tuned models being served. Since experts are the major contributor to the model's large size, we propose to find the most similar experts and load them as common experts in a round-robin approach to reduce the overall memory footprint.
    \item Runtime Partial Reconfiguration: Despite having less number of parameters compared to experts, The contribution of non-expert layers is paramount in forward pass calculation. Therefore, to serve requests for a certain model, we instantly reconfigure the system by only swapping the non-expert layers loaded on GPU with non-expert layers of the requested model.
    \item We implement and deploy our serving approach on a system equipped with a single NVIDIA A100 GPU. To evaluate the impact of our method on generated output quality, we benchmark two model families—Mixtral-8x7B and Google Switch Transformer Base-8 on tasks such as language generation and instruction following. Additionally, we use a Poisson distribution with varying arrival rates to assess the quality of service (QoS) provided by our approach, and compare it against NVIDIA MIG as a baseline.
\end{itemize}

\section{System Overview and Problem Statement}

\vspace{-0.8cm}

\begin{figure}[ht]
\vskip 0.2in
\begin{center}
\centerline{\includegraphics[width=0.9\columnwidth]{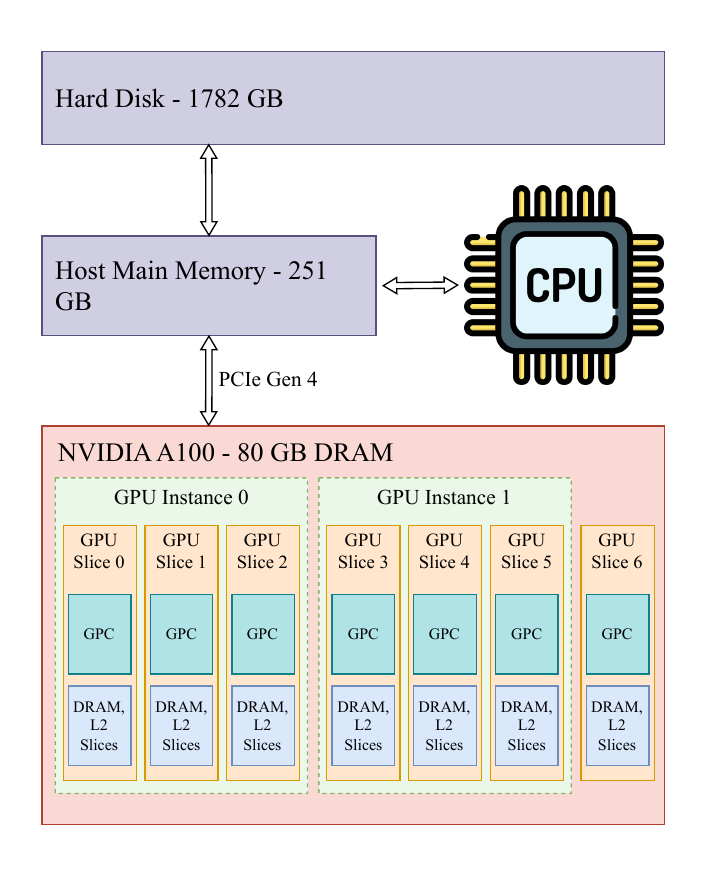}}
\caption{Diagram of an inference system with a single NVIDIA A100 GPU.}
\label{system_overview}
\end{center}
\vskip -0.2in
\end{figure}

\begin{figure*}[ht]
\vskip 0.2in
\begin{center}
\centerline{\includegraphics[width=1.8\columnwidth]{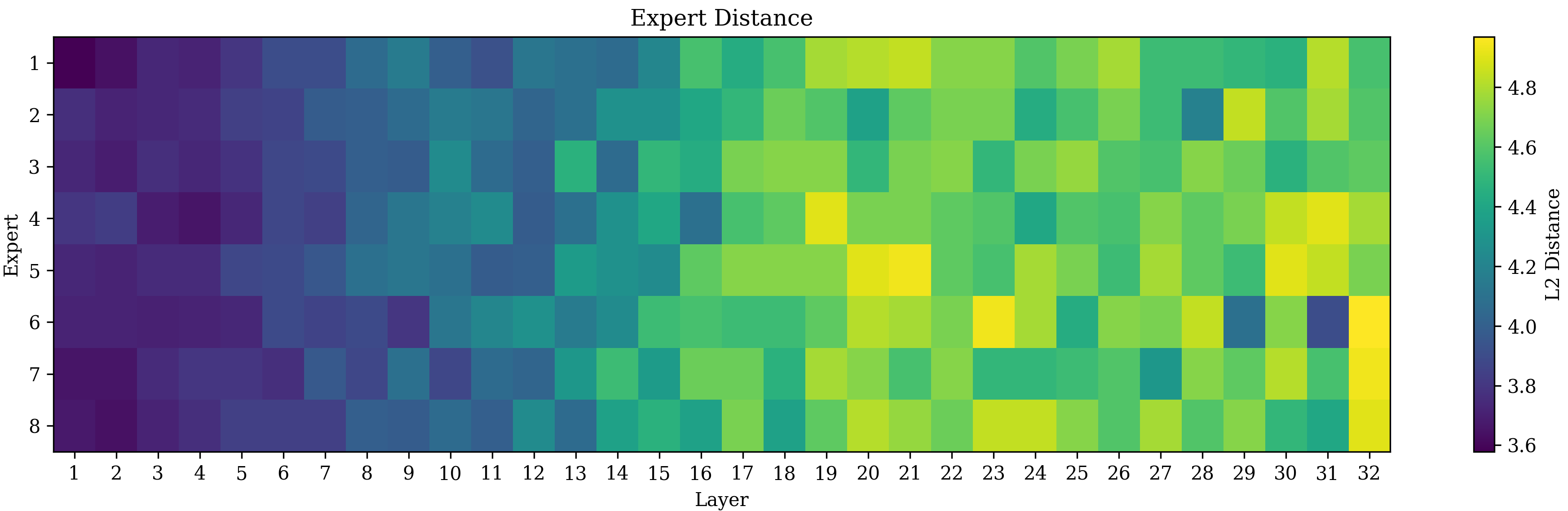}}
\caption{Expert-to-expert distance between Mixtral base and instruct model: Despite being fine-tuned for different tasks, experts in the same positions exhibit similarity.}
\label{expert_distance}
\end{center}
\vskip -0.2in
\end{figure*}

The components of a single-GPU inference system are depicted in \cref{system_overview}. In this single-GPU system, the models are originally stored in the hard disk. For every ML serving application, the parameters must first be loaded into the host's main memory. If the inference is to be performed on the accelerator, the model parameters will be transferred from the host's main memory to the GPU's memory. After the load is completed, the model will be ready for inference, and for every new request, the inputs are copied to GPU, processed and the results will be written back to CPU's main memory. 

However, when handling an MoE LLM, the model does not completely fit in the accelerator's memory. Originally proposed in \cite{eliseev2023fast}, offloading techniques load the non-expert parameters of the model to GPU initially and fill the remaining space with as many experts as possible. When performing inference for a sparse MoE model, if the required expert is not available on GPU's memory, it must be transferred from the CPU's memory through the PCIe link, which stalls the inference process and slows the token generation speed.

In this paper, we consider a single-GPU inference system that has to support two or more MoE LLMs. For such a problem, accelerator resources must be shared and the conventional virtualization approaches will be considered as follows:

\textbf{Time Sharing.} For this approach, there will be a queue that handles requests coming from the users. If the next request's requested model matches with the currently loaded model, the inference will be performed. However, in a case of miss-match, the whole model has to be offloaded to CPU's main memory and the requested model should be uploaded onto GPU. Due to the large size of the discussed models and based on our experiments, each model swap can take up to 2 minutes, which will inject a considerable delay in the inference process.

\textbf{Space Sharing.} For space-sharing the resources on GPU should be split among the models we are trying to serve. The A100 NVIDIA GPU consists of seven GPU slices and starting with NVIDIA Ampere architecture each slice is capable of operating independently. Each slice comes with its own memory (DRAM and L2 slices), compute resource GPU Processing Cluster (GPC), and host-to-device link bandwidth. On an A100, each GPC is consisted of 7 Texture Processing Clusters (TPC), and each TPC is comprised of 2 streaming multiprocessors (SMs). NVIDIA MIG \cite{choquette2021nvidia} allows us to make different partitions using different numbers of GPU slices. For example, in \cref{system_overview}, the system is partitioned into two equal GPU instances which can support the inference of two models simultaneously and independently.

\vspace{-0.3cm}

\section{Approach}

\subsection{Parameter Layout on GPU: A Consolidated Middle-Point}

In the described environment, the parameters of the MoE model can be grouped into three categories: on-device non-expert weights, on-device expert weights, and off-device expert weights. The on-device expert weights constitute the majority of the parameters loaded into GPU memory. Accordingly, in this section, we present the algorithm that generates a unified layout for the on-device expert weights, which is used during the initial model load.

As shown in \cref{expert_distance}, the expert-to-expert distances between a fine-tuned instruct model and a base model are not equal. In this context, we flatten the weights of an expert block into a 1D tensor and define the distance as the $L2 \;distance$ between experts from different models. As depicted, the distance between experts generally increases in the deeper layers, which aligns with the experiments conducted in \cite{shen2024efficient}. While the distances between same expert positions are in the range of 3.6 to 4.9, expert-to-expert distances with different layer and expert numbers can vary from 150 to 250. This observation leads us to propose using the experts interchangeably.

Therefore, to find the most similar experts and generate a unified layout, we associate a degree of similarity to each expert coordinate specified by layer and expert numbers. In the case of having only two models for serving, the degree of similarity for each location is the expert-to-expert distance for a given $(layer, expert)$ pair. However, for higher number of models, the distance for each combination will be the sum of all model-to-model combinations.

Subsequently, each expert location will be ranked based on the defined similarity measure, with the highest rank indicating the greatest similarity, corresponding to the lowest distance. To generate the initial loader's expert map, starting from the highest rank, each expert location will be assigned to a specific fine-tuned model using a round-robin approach. For instance, if we are serving two models, the expert locations with odd ranks will be assigned to the first model, while those with even ranks will be assigned to the second model. Although the time spent generating this expert map can be substantial and depends on the number of models, this process only needs to be performed once for each combination prior to load time and can be done offline.

\begin{algorithm}[tb]
   \caption{Initial Expert Loader}
   \label{loader}
   \begin{algorithmic}[1]
       \Statex {\bfseries Input:} List of models $\mathcal{M}$, expert capacity $C$

       \Statex \textcolor{gray}{\small \#Iterate over layers and experts}
       \Statex \textbf{for} {$(iL, iE) \in \{1,\cdots, L\}\times\{1, \cdots, E\}$} \textbf{do}  
               \Statex \hspace{0.5cm} \textcolor{gray}{\small \#Iterate over all model pairs}
               \Statex \hspace{0.5cm} \textbf{for} {$(i, j) \in \mathcal{M} \times \mathcal{M}$} \textbf{do} 
                    \Statex \hspace{1.0cm} $\texttt{distance}[iL][iE] \gets  \texttt{distance}[iL][iE] +$
                    \Statex \hspace{0.25cm} $\|\mathcal{M}[i].\texttt{expert}[iL][iE] - \mathcal{M}[j].\texttt{expert}[iL][iE]\|_2$

            \Statex \hspace{0.25cm} \textbf{end for}
        \Statex \textbf{end for}
       \Statex \textcolor{gray}{\small \#Initialize counter for expert assignments}
       \Statex Initialize \texttt{count} $\gets 0$

       \Statex \textcolor{gray}{\small \#Assign experts based on sorted similarity}
       \Statex \textbf{for} {$(iL, iE)$ in \texttt{sorted(distance)} \textbf{while} \texttt{count} $< C$} \textbf{do}
           \Statex \hspace{0.5cm} \texttt{count} $\gets$ \texttt{count} + 1
           \Statex \hspace{0.5cm} $idx \gets \texttt{count} \mod |\mathcal{M}|$
           \Statex \hspace{0.5cm} $\texttt{map}[iL][iE] \gets \mathcal{M}[idx].\texttt{id}$
           \Statex \hspace{0.5cm} Load $\mathcal{M}[idx].\texttt{expert}[iL][iE]$ to GPU
        \Statex \textbf{end for}

       \Statex \textcolor{gray}{\small \#Load non-expert model weights}
       \Statex Set \texttt{loadedModel} to the first model id in $\mathcal{M}$
       \Statex Load non-expert weights of \texttt{loadedModel} to GPU
   \end{algorithmic}
\end{algorithm}

This approach sets the granularity of model merging at the expert level and ensures that the loaded parameters are equally representative of all the models being served. After generating the expert map, the parameters should be loaded to the GPU memory. Based on the available GPU memory and the size of non-expert parameters, the total number of experts that can fit on the GPU will be calculated. To start, non-expert parameters of a random model from our model list will be loaded to GPU's memory. Following this, the loader begins with the highest-ranked expert locations, and transfers their corresponding parameters from the host's main memory to the GPU's memory.

\begin{algorithm}[h!]
    \caption{Inference Orchestrator}
    \label{orchestrator}
    \begin{algorithmic}[1]
        \Statex {\bfseries Input:} prompt \texttt{prompt}, requested model \texttt{targetModel}
        \Statex \textbf{If} {\texttt{targetModel} is not \texttt{loadedModel}} \textbf{then}
            \Statex \hspace{0.5cm} Load \texttt{targetModel} non-experts on GPU
        \Statex \textbf{EndIf}

        \Statex \textcolor{gray}{\small \#Tokenize the input prompt and embed it into activations}
        \Statex \textbf{Repeat}
            \Statex \hspace{0.25cm} \texttt{act} =  tokenize and embed \texttt{prompt}.
            \Statex \hspace{0.25cm} \textcolor{gray}{\small \#Process the token through all transformer layers}
            \Statex \hspace{0.25cm} \textbf{For} {$iL = 1$  to $L$} \textbf{do}
                \Statex \hspace{0.5cm} \textcolor{gray}{\small \#Apply the multi-headed attention}
                \Statex \hspace{0.5cm} \texttt{act} =
                \Statex \hspace{0.6cm}\texttt{layers[$iL$].MultiheadAttention(act)}

                \Statex \hspace{0.5cm} \textcolor{gray}{\small \#Determine which expert to use}
                \Statex \hspace{0.5cm} $(iL, iE)$ = \texttt{layers[$iL$].gate(act)}

                \Statex \hspace{0.5cm} \textbf{If} \texttt{map[$iL$][$iE$] is not on GPU} \textbf{then}
                    \Statex \hspace{0.75cm} \textcolor{gray}{\small \#Load expert from CPU if not on GPU}
                    \Statex \hspace{0.75cm} Load \texttt{targetModel.expert[$iL$][$iE$]} to 
                    \Statex \hspace{0.75cm} GPU
                    \Statex \hspace{0.75cm} \texttt{act} =
                    \Statex \hspace{1cm} \texttt{targetModel.expert[$iL$][$iE$](act)}
                \Statex \hspace{0.5cm} \textbf{Else}
                    \Statex \hspace{0.75cm} \textcolor{gray}{\small \#Use the preloaded expert}
                    \Statex \hspace{0.75cm} $id$ = \texttt{map[$iL$][$iE$]}
                    \Statex \hspace{0.75cm} \texttt{act} = \texttt{$\mathcal{M}$[$id$].expert[$iL$][$iE$](act)}
                \Statex \hspace{0.5cm} \textbf{EndIf}
                \Statex \hspace{0.5cm} Perform Add \& Norm and generate \texttt{act} of next 
                \Statex \hspace{0.5cm} layer
            \Statex \hspace{0.25cm} \textbf{EndFor}

            \Statex \hspace{0.25cm} Predict \texttt{nextToken} using \texttt{act}

            \Statex \hspace{0.25cm} Concatenate \texttt{nextToken} to the end of \texttt{prompt}
        \Statex \textbf{Until} {\texttt{nextToken} is \texttt{eos} or required number of output tokens generated}
    \end{algorithmic}
\end{algorithm}

\vspace{-0.3cm}



\begin{figure*}[ht]
\vskip 0.2in
\begin{center}
\centerline{\includegraphics[width=1.9\columnwidth]{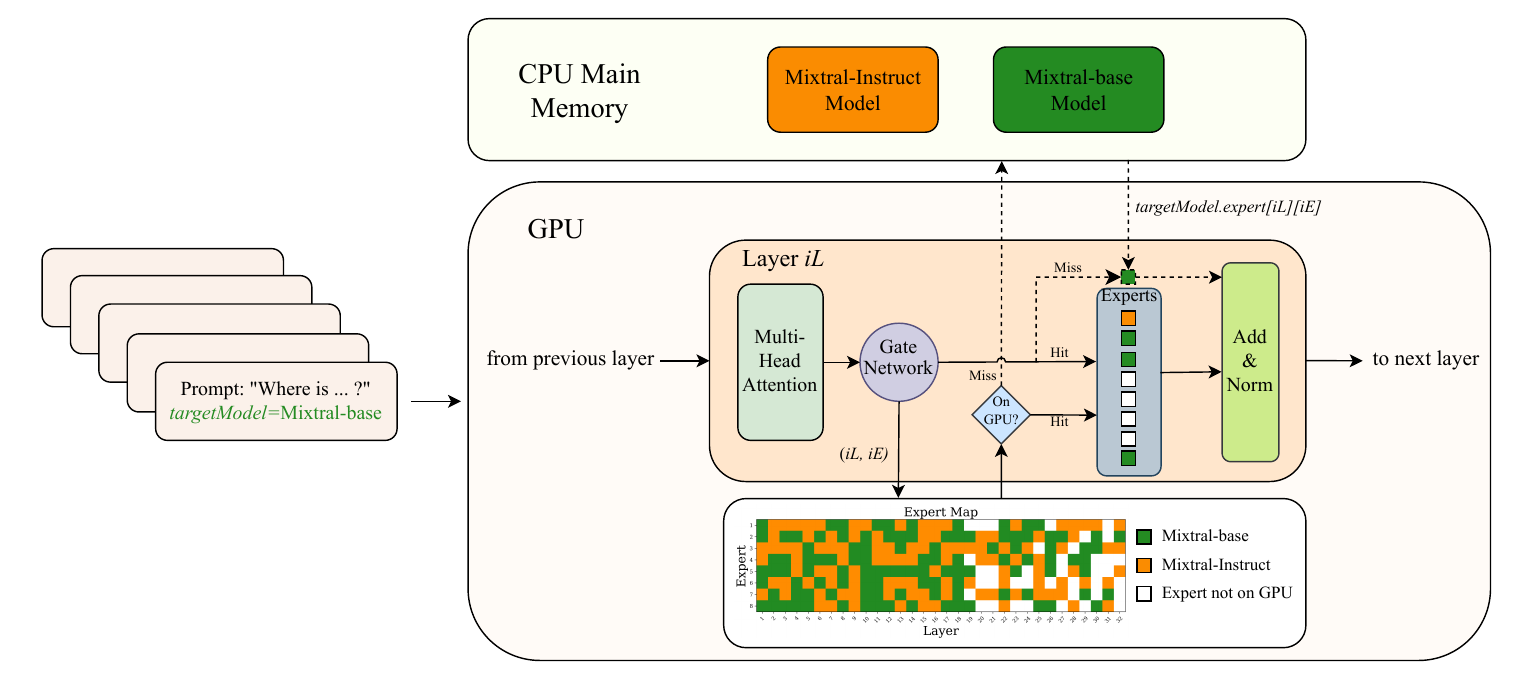}}
\caption{Inference process for each layer of the consolidated model: In the case of an expert hit, the inference is performed using the already loaded expert, which may not necessarily belong to the requested model. However, in the case of a miss, the corresponding expert is loaded from the host's memory to preserve the quality of the generated output.}
\label{layerinference}
\end{center}
\vskip -0.2in
\end{figure*}

As summarized in \cref{loader}, the loader's inputs are a list of models being served and the total expert capacity. At the beginning of the loading process, all the models' weights are stored on CPU's main memory. By the end of the process, a fraction of each model's parameters is loaded onto the GPU, collectively forming a unified model ready for inference. Moreover, the generated $map$ is a table that keeps track of each expert and its corresponding model id which will route each input in performing forward pass calculations.

\vspace{-0.3cm}

\subsection{Inference Orchestration}

The loader, as explained in \cref{loader}, loads experts from different models. While this approach significantly reduces the memory footprint required to serve all the models, it may decrease the quality of output generation for tasks specific to each model variant.

As an example, the Mixtral model evaluated in this paper contains a total of 1,605.64 million non-expert parameters. Its architecture comprises 32 layers, each with 8 experts, and utilizes top-2 sparse routing in the gate network. During each forward pass, a total of 1,605.64 million non-expert parameters and $32 \times 2 \times 176.16$ million expert parameters (with each expert contributing 176.16 million parameters) are actively involved in computations. Consequently, the non-expert parameters contribute approximately 14.2\% to the generated output.

In contrast to experts, non-expert parameters are always utilized in each forward pass, regardless of activations. Furthermore, their total size is approximately 3 GB (2 bytes per parameter). Based on the described system, it takes approximately a second to transfer all non-expert parameters of a given model via the host-to-device PCIe link. Consequently, for each new request to the serving system, the $targetModel$ non-expert parameters are transferred to the GPU. Although this process incurs a one-time cost per request and increases the TTFT, it ultimately enhances the quality of the generated output.

Accordingly, the inference orchestrator begins handling each request by comparing the $targetModel$ with the $loadedModel$. If they match, inference is performed immediately. However, if they do not match, the $targetModel$ non-expert parameters are uploaded to the GPU, replacing the currently loaded non-expert parameters before proceeding with inference.

As summarized in \cref{orchestrator}, during the inference phase, each sequence undergoes tokenization and embedding processes before arriving at the first layer of the model. Within each layer, the activations are first processed by a multi-headed attention layer and then passed to the gate network. The gate network generates weighted averaging coefficients for each expert, and in a sparse gating scenario, the top experts are selected to produce the output of the layer.

As depicted in \cref{layerinference}, pairs of layer and expert numbers ($(iL, iE)$) are generated, and the status of the requested expert is retrieved from the expert map. If the expert is already loaded on the GPU, the corresponding expert's result will be calculated, although it is not guaranteed to belong to the $targetModel$. However, in the case of an expert miss (dashed lines in \cref{layerinference}), a load request is sent to the CPU's main memory. Since all models are stored in the CPU's main memory, the requested expert will belong to the $targetModel$, preventing further quality degradation. Finally, after processing all layers, the activations generated by the last layer pass through the \textit{language model head} to produce the output logits.

\vspace{-0.2cm}
\section{Evaluation Methodology}

We evaluate our serving system using the Mixtral-8x7B-v0.1 and Google Switch Transformer Base-8 family of models. The Mixtral model has 32 layers and 8 experts per layer and comes in two variants. The Switch Transformer model comes in four variants and has a total of 96 experts, which are distributed equally across 12 layers (8 experts per layer). Since in our design, both QoS and the application output quality is affected, we evaluate the proposed system through experiments that independently assess each aspect of its functionality.

\vspace{-0.25 cm}
\subsection{Quality of Service}

Since the Switch Transformer Base-8 model has a considerably lower memory footprint (1.5 GB) compared to Mixtral and does not constrain our environment, we use only the Mixtral model and its variants for this experiment. We evaluate our proposed system in a scenario where requests arrive independently for each model we support. We consider two separate Poisson processes to generate requests for each supported model. To demonstrate the system's responsiveness and performance under varying loads, we evaluate the system at different arrival rates ($\lambda$). Each request to the system is a prompt consisting of 20 tokens, with the number of requested output tokens set to 25. To compare the QoS of each approach, we measure the average TTFT, average turnaround time, and the total number of processed requests (throughput).

We implement our loader and orchestrator using the PyTorch \cite{paszke2019pytorch} framework. Our experiments are conducted on a server with an AMD 16-Core MILAN CPU and a single 80GB NVIDIA A100 GPU, connected via a PCIe Gen4 host-to-device interconnect, as outlined in the system overview section. QoS metrics are measured from PyTorch’s perspective using Python’s time package. It is important to note that other inference optimizations, such as Flash Attention \cite{dao2022flashattentionfastmemoryefficientexact}, are complementary to the approach presented in this paper and fall outside the scope of this work. To evaluate the system’s performance, we focus on single-batch requests, providing a controlled setting for our experiments.

\textbf{QoS Baselines.} 
We compare our serving system against two baselines. The first baseline represents a scenario in which a single Mixtral model is served, but with an arrival rate that is twice that of each Poisson process used to evaluate the proposed approach. The second baseline involves partitioning the GPU resources into two using NVIDIA MIG, with each partition assigned to an independent process that handles requests for the base and instruct models, separately.

\vspace{-0.25 cm}
\subsection{Application Quality}

The designed serving system must provide accessibility to multiple models. Therefore, for the Mixtral model, we measure the quality of the generated output using two specific families of benchmarks corresponding to each model variant: the Base model and the Instructed model. During the evaluation of each benchmark family, the non-expert layers of the model are set to the respective model variant.

For the Base model, we evaluate the quality of text generation using the WikiText2 \cite{merity2016pointer}, PTB \cite{marcus1994penn}, and C4 \cite{raffel2020exploring} datasets. For each dataset, we assess the models' generated output by measuring perplexity using 128 samples, with each sample consisting of 2048 tokens.

To demonstrate the quality of output for instruct tasks, we evaluate each model using the MT-Bench \cite{zheng2023judging} benchmarks. MT-Bench is a benchmark specifically designed for instruct-based models and includes 80 instructions across 8 different categories. Each request prompts the language model to generate an output, and requests may include a follow-up instruction to refine the output and further improve its quality. The quality of the responses is then judged by the GPT-4 \cite{achiam2023gpt} model, which assigns each response a grade from 0 to 10. MMLU, HellaSwag, and TruthfulQA are multiple-choice question benchmarks that assess the contextual understanding and reasoning abilities of language models. 

Moreover, to demonstrate the general effectiveness, we asses the performance of each model variant using HellaSwag \cite{zellers2019hellaswag}, MMLU \cite{hendrycks2020measuring}, and TruthfulQA \cite{lin2021truthfulqa} benchmarks. For MMLU and TruthfulQA, we use a 5-shot prompt, and for HellaSwag, we use a 0-shot prompt. For MMLU and HellaSwag, we evaluate each model with 1000 random samples, and for TruthfulQA, we evaluate it with 812 samples.

\textbf{Baselines.} We compare the quality of the generated output from the proposed consolidated model against the following models: Mixtral-8x7B-v0.1-base, Mixtral-8x7B-v0.1-Instruct, and Mixtral-8x7B-v0.1-avg. The first two models are official releases by Mistral-AI. However, proposed in \cite{izmailov2018averaging}, the Mixtral-8x7B-v0.1-avg model is generated by averaging the weights of the first two models. We include Mixtral-8x7B-v0.1-avg in the comparison to evaluate the performance of the proposed model against conventional model-merging approaches. It is important to note that for benchmarks measuring perplexity, the non-expert parameters in our approach come from the base model, while for all other benchmarks, they are derived from the instruct model.

\subsection{Scalability}

To demonstrate the scalability and applicability of our approach to other model architectures with a higher number of fine-tuned variants, we evaluated our approach against the Averaging baseline \cite{izmailov2018averaging} using Google’s Switch Transformer Base-8 family of models. For this experiment, we use the base model provided by Google along with three other community-provided fine-tuned variants: 

\begin{itemize}
    \item Model A: \texttt{emre/switch-base-8 \\ -finetuned-samsum}
    \item Model B: \texttt{google/switch-base-8}
    \item Model C: \texttt{glamprou/switch-base-8-sst2}
    \item Model D: \texttt{glamprou/switch-base-8-mnli}
\end{itemize}


To evaluate how output quality is affected by the number of merged models, we consider three serving scenarios: 

\vspace{-0.3cm}
\begin{itemize}
    \item Serving models A \& B
    \item Serving models A \& B \& C
    \item Serving models A \& B \& C \& D
\end{itemize}

\begin{figure*}[ht]
\vskip 0.2in
\begin{center}
\centerline{\includegraphics[width=1.9\columnwidth]{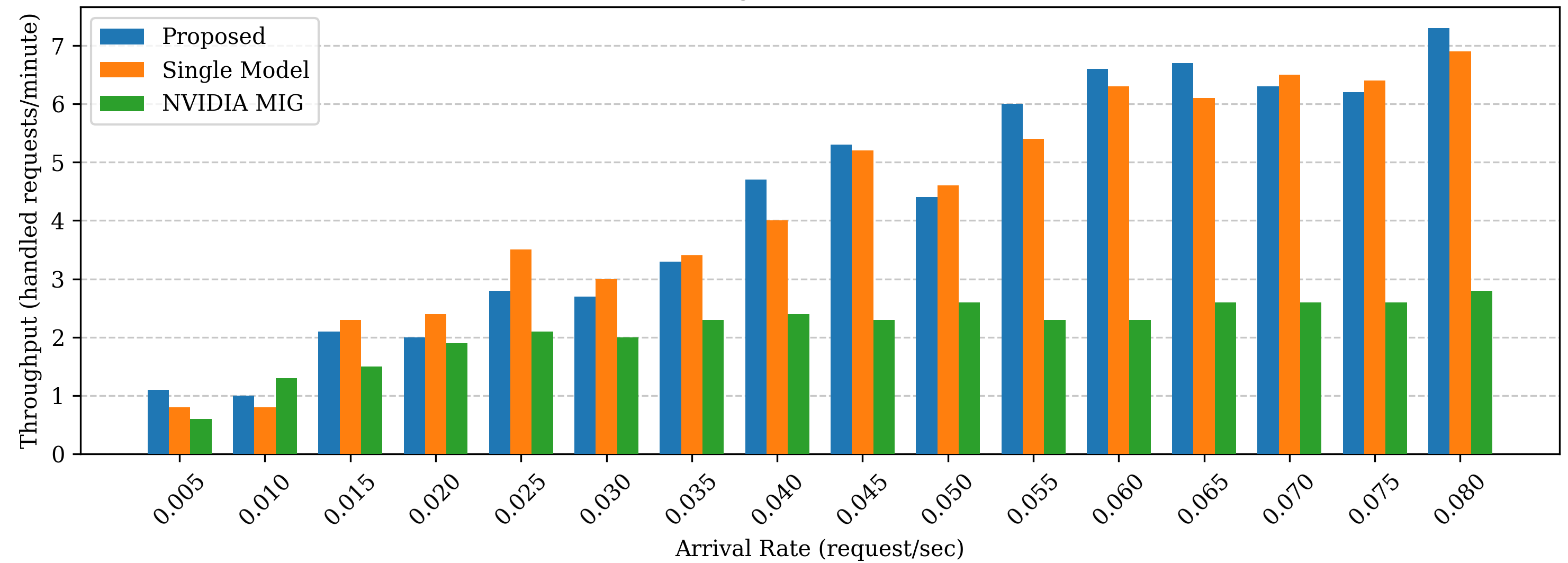}}
\caption{The throughput of each serving approach is measured in completed requests per minute. Since NVIDIA MIG utilizes two independent processes to handle requests for different models, the arrival rate for each instance is half of the value shown on the x-axis, and the reported throughput is the sum of the throughput for both instances. The reported values represent the average over five independent runs.}
\label{throughput}

\end{center}
\vskip -0.2in
\end{figure*}

\vspace{-0.4cm}

\section{Results}

\subsection{Performance and Quality of Service}

\cref{throughput} compares the throughput of each serving approach across varying arrival rates. For NVIDIA MIG, the reported throughput represents the combined throughput of two independent GPU instances, as illustrated in \cref{system_overview}. At lower arrival rates, all three approaches exhibit low throughput, which is attributed to system under-utilization, meaning they can handle all incoming requests within the experiment's time window. As the arrival rate increases, throughput improves until reaching distinct ridge points for each approach. For both the proposed system and the single-model serving approach, the ridge point occurs at $\lambda = 0.060$, while for NVIDIA MIG, it is observed at $\lambda = 0.040$. This indicates that the proposed system and the single-model approach have higher capacities and can continue handling more requests at higher arrival rates, whereas NVIDIA MIG's performance plateaus earlier.

\cref{ttft_turnaround} illustrates the average TTFT and turnaround times, both normalized to their maximum values. Single model serving and the proposed system achieve significantly lower TTFT (a maximum of 5.86 seconds) and turnaround times (a maximum of 49.67 seconds) compared to NVIDIA MIG, which supports the results depicted in \cref{throughput}. As demonstrated, each NVIDIA MIG instance takes an average of 49 seconds to complete a request.

Moreover, comparing the proposed system against the single-serving approach highlights the overhead of non-expert runtime reconfiguration. As shown in the TTFT section of \cref{ttft_turnaround}, our approach introduces a delay of approximately half a second on average. However, the actual measured latency for swapping non-experts is approximately 1.2 seconds, which does not occur for consecutive requests targeting the same $targetModel$. It is also important to note that the numbers reported in this subsection are specific to the requests defined earlier and may vary for prompts with different lengths and requested numbers of output tokens.


To provide deeper insights into the performance of each approach, the inference latency for each model layer is presented in \cref{layer_latency}. As mentioned in \cite{scaling-book}, during the generation phase of a transformer-based LLM, the attention kernel is memory-bound. This is because the results of prior computations, stored in the KV cache, must be copied from the GPU’s GMEM to SMEM. Consequently, each GPU instance in NVIDIA MIG has sufficient compute resources, and we do not observe a significant increase in the latency of the attention layer (0.72 -> 0.78).

However, the expert block is compute-bound \cite{scaling-book}. Therefore, when both required experts are available in GMEM, the latency increase is more noticeable (1.2 -> 1.7). On the other hand, when the hit rate decreases for the expert block, an overhead (27.8 ms for each expert) is introduced due to copying experts from the CPU’s DRAM to the GPU’s GMEM via the PCIe link. In our approach, when serving a single model, the full bandwidth of the PCIe link is available to the process. However, using NVIDIA MIG with two GPU instances splits the available PCIe bandwidth between them, nearly doubling the imposed overhead (51.3 ms for each expert).

\begin{table}[]
\caption{Latency (milliseconds) of a layer for different hit rates (out of 2).}
\vskip 0.1in 
\label{layer_latency}
{\renewcommand{\arraystretch}{1.5}

\small
\begin{tabular}{ccccc}
\hline
\multirow{2}{*}{}                                          & \multirow{2}{*}{Attention} & \multicolumn{3}{c}{Hit Rate}    \\ \cline{3-5} 
                                                           &                            & 0 expert & 1 expert & 2 experts \\ \hline
\begin{tabular}[c]{@{}c@{}}Single/\\ Proposed\end{tabular} & 0.72                       & 56.8     & 29.2     & 1.2       \\
NVIDIA MIG                                                 & 0.78                       & 104.3    & 54.1     & 1.7       \\ \hline
\end{tabular}
}
\end{table}

\begin{table}[]
\centering
\caption{Average TTFT and turnaround time (seconds) for the proposed and compared serving approaches.}
\vskip 0.1in 
\label{ttft_turnaround}
{\renewcommand{\arraystretch}{1.5}
\small
\begin{tabular}{clclc}
\hline
           &  & TTFT &  & Turnaround Time \\ \hline
Single     &  & 0.89 &  & 8.34       \\
Proposed   &  & 1.41 &  & 8.78       \\
NVIDIA MIG &  & 5.86 &  & 49.67      \\ \hline
\end{tabular}
}
\end{table}

\subsection{Quality of Generated Output}

\begin{table*}[]
\centering
\caption{The quality of the generated output, evaluated across different task families, including language generation (Perplexity), instruction following (MT-Bench), and general-purpose multiple-choice questions.}
\vskip 0.1in 
\label{quality_comparison}
{\renewcommand{\arraystretch}{1.6}
\small
\begin{tabular}{ccccccccccc}
\hline
\multirow{2}{*}{Model}              & WikiText $\downarrow$      & C4 $\downarrow$            & PTB $\downarrow$            & \multirow{2}{*}{} & \multicolumn{3}{c}{MT-Bench (out of 10) $\uparrow$}      & \multirow{2}{*}{\begin{tabular}[c]{@{}c@{}}MMLU $\uparrow$\\ (5-shot)\end{tabular}} & \multirow{2}{*}{\begin{tabular}[c]{@{}c@{}}HellaSwag $\uparrow$\\ (0-shot)\end{tabular}} & \multirow{2}{*}{\begin{tabular}[c]{@{}c@{}}TruthfulQA $\uparrow$\\ (5-shot)\end{tabular}} \\ \cline{2-4} \cline{6-8}
                                    & \multicolumn{3}{c}{(Perplexity)}               &                   & 1st Turn      & 2nd Turn      & Avg           &                                                                          &                                                                               &                                                                                \\ \cline{1-5} \cline{6-9} \cline{10-11} 
Base              & 3.81          & 7.24          & 13.59          &                   & 3.01          & 2.16          & 2.60          & 70.0\%                                                                   & 81.1\%                                                                        & 66.6\%                                                                         \\
Instruct          & 4.12          & 7.62          & 14.60          &                   & 8.28          & 7.98          & 8.13          & 71.2\%                                                                   & 83.0\%                                                                        & 73.6\%                                                                         \\
Avg               & 3.89          & 7.38          & 13.44          &                   & 8.38          & 7.61          & 7.99          & 72.2\%                                                                   & 82.0\%                                                                        & 73.0\%                                                                         \\
\textbf{Proposed} & \textbf{3.85} & \textbf{7.33} & \textbf{13.90} &                   & \textbf{8.33} & \textbf{7.98} & \textbf{8.16} & \textbf{71.7\%}                                                                   & \textbf{82.2\%}                                                               & \textbf{70.6\%}                                                                \\ \hline
\end{tabular}
}
\end{table*}

\cref{quality_comparison} presents a comparative analysis of the generated output quality across various models, evaluating perplexity on the WikiText, C4, and PTB datasets, as well as performance metrics on MT-Bench, MMLU, HellaSwag, and TruthfulQA benchmarks. Lower perplexity scores indicate better language modeling capabilities, while higher scores indicate improved performance on the remaining tasks. While the base model serves as the optimal baseline for text generation, the proposed approach outperforms the averaged model and achieves lower perplexity across all datasets, except for PTB, which is shown to exhibit irregular performance improvements even at higher levels of quantization \cite{eliseev2023fast}.

On MT-Bench, the proposed model outperforms others with an average score of 8.16, closely aligning with the high-quality outputs of the instruct-tuned model. Notably, although the averaged model demonstrates superior performance in the first turn, its quality drops significantly when responding to the follow-up refinement prompt and exhibits high variance.

Additionally, the proposed system achieves strong results on MMLU (71.7\%), HellaSwag (82.2\%), and TruthfulQA (70.6\%) which demonstrates a balanced performance. These results highlight the ability of the proposed system to maintain high output quality at other general-purpose benchmarks.

\vspace{-0.2cm}
\subsection{Scalability}

\begin{figure*}[ht]
\vskip 0.2in
\begin{center}
\centerline{\includegraphics[width=2\columnwidth]{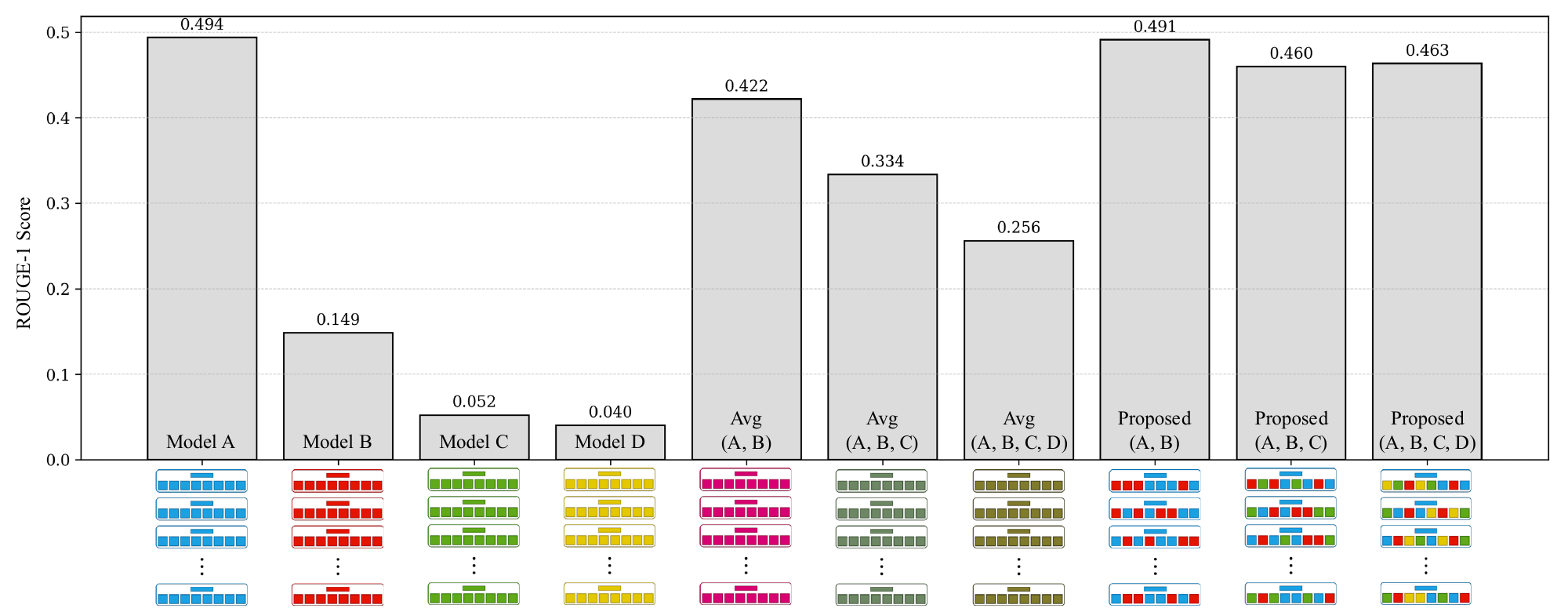}}
\caption{ROUGE-1 scores (higher is better) on the SAMSum dataset for individual models and their combinations, comparing the proposed approach with the model-merging baseline. Model layouts are illustrated beneath each bar. In our approach, a portion of the expert parameters (squares) and all non-expert parameters (rectangles) are retained from the model specifically trained for the benchmark (Model A, shown in blue). In contrast, the averaging approach modifies all parameters by averaging across the merged models.}
\label{scalability}

\end{center}
\vskip -0.2in
\end{figure*}

Presented in \cref{scalability}, we report the ROUGE-1 scores (an N-gram based summarization evaluation metric) for a summarization task on the SAMSum dataset \cite{Gliwa_2019}, which Model A is specifically finetuned for. As shown, increasing the number of merged models significantly degrades output quality for the Averaging baseline, with ROUGE-1 scores dropping from 0.49~$\rightarrow$~0.42~$\rightarrow$~0.33~$\rightarrow$~0.25. However, our approach demonstrates greater resilience, maintaining higher output quality even with more variants, with scores of 0.49~$\rightarrow$~0.49~$\rightarrow$~0.46~$\rightarrow$~0.46. For detailed results, including ROUGE-2, ROUGE-L, and ROUGE-Lsum—which follow a similar trend—please refer to \cref{samsum} in \cref{app_A}.

\vspace{-0.25 cm}

\section{Related Work}

Reducing the overhead of the host-to-device link is crucial for efficiently serving MoE models in constrained environments and efficient expert scheduling and model compression techniques such as pruning, and quantization \cite{lin2023awq, huang2024billm, eliseev2023fast, imani2024mixture, dettmers2023case, xiao2023smoothquant} have been utilized extensively \cite{liu2024survey} to speed up the inference process.

\textbf{Model Pruning    } 
Expert pruning in MoE models aims to reduce the number of parameters while maintaining model accuracy. This process is typically categorized into structured and unstructured pruning. Structured pruning methods focus on reducing the number of experts, with some approaches directly removing unimportant experts and others merging them. Importance of an expert can be defined using different metrics. For example, approaches proposed in \cite{chen2022task, muzio2024seer, park2024learning} identify the unimportant experts based on frequency and scarcity and prunes non-essential experts while fine-tuning important ones for target tasks. Moreover, similar to our approach, authors in \cite{chowdhury2024provably, yang2024moe} propose to prune the experts of a fine-tuned model that have lower l2 distances from the base model. However, these approaches focus on performance of a single models and do not consider a multi-model multi-task system.

\textbf{Multi-Task Model Merging    }
Model merging integrates multiple models into a single model by performing weight interpolation at the parameter level, serving as an efficient alternative \cite{singh2020model, yang2024surgeryv2, yang2023adamerging}.  Existing model merging methods include weighted merging \cite{matena2022merging, wortsman2022model, jin2022dataless}, which assigns varying importance to different models, and subspace merging \cite{du2024parameter, yadav2023resolving}, which eliminates unimportant neurons to reduce task interference. However, these approaches use static merging strategies, limiting adaptability. Moreover, techniques such as activation matching, and permutation invariance help minimize discrepancies. Although these approaches demonstrate improved performance on considerably smaller models, their applicability for larger MoE architectures remains unexplored. Additionally, unlike previous methods that rely solely on static merging, our approach enhances both efficiency and effectiveness by integrating dynamic runtime reconfigurability with static merging.

\vspace{-0.25 cm}
\section{Conclusion}

In this paper, we presented an efficient serving system designed to address the challenges of deploying multiple fine-tuned MoE-based LLMs in single-GPU resource-constrained settings. By leveraging similarity-based consolidation, our approach reduced the overall memory footprint by sharing similar experts across models, and runtime partial reconfiguration ensured that non-expert layers are dynamically swapped to maintain competitive output quality. Our evaluation, conducted on a server with a single NVIDIA A100 GPU, demonstrated the system's ability to significantly reduce turnaround time compared to conventional virtualization approaches like NVIDIA MIG which resulted in achieving a comparable throughput to single model serving, with only a minimal increase in TTFT. The results also highlighted that our approach maintained high output quality, as evidenced by strong performance across various benchmarks.

\section*{Acknowledgements}

This work was supported by the National Science Foundation under Grant No. 2038682. We thank Pedram Akbarian, and Rasool Sharifi for the helpful discussions.

\bibliography{references}
\bibliographystyle{icml2025}

\newpage
\appendix
\onecolumn
\section{Complete Results of SAMSum Dataset}
\label{app_A}

\cref{samsum} presents the detailed results for the summarization task on the SAMSum dataset. As demonstrated, all metrics—including ROUGE-2, ROUGE-L, and ROUGE-Lsum—follow a similar trend, highlighting the greater resilience of our approach as the number of served models increases.

\begin{table}[H]
\centering
\caption{ROUGE scores (higher is better) on the SAMSum dataset for individual models and their combinations.}
\label{samsum}
{\renewcommand{\arraystretch}{1.5}
\begin{tabular}{ccccc}
\hline
Model                                                & ROUGE-1 & ROUGE-2 & ROUGE-L & ROUGE-Lsum \\ \hline
Model A                                              & 0.493   & 0.250   & 0.409   & 0.409      \\
Model B                                              & 0.148   & 0.026   & 0.128   & 0.128      \\
Model C                                              & 0.052   & 0.004   & 0.046   & 0.046      \\
Model D                                              & 0.040   & 0.009   & 0.036   & 0.036      \\
Avg (A, B)                                           & 0.422   & 0.201   & 0.344   & 0.344      \\
Avg (A, B, C)                                        & 0.333   & 0.132   & 0.277   & 0.277      \\
Avg (A, B, C, D)                                     & 0.256   & 0.079   & 0.209   & 0.209      \\
Proposed (A, B)                                      & 0.491   & 0.245   & 0.407   & 0.406      \\
Proposed (A, B, C)                                   & 0.460   & 0.229   & 0.377   & 0.377      \\
Proposed (A, B, C, D)                                & 0.463   & 0.229   & 0.377   & 0.377      \\ \hline
\end{tabular}
}
\end{table}

\end{document}